\documentclass{article}

\usepackage[preprint,nonatbib]{neurips_2021}

\usepackage{graphicx}
\usepackage{mathtools}
\usepackage{url}            % simple URL typesetting
\usepackage{booktabs}       % professional-quality tables
\usepackage{amsfonts}       % blackboard math symbols
\usepackage{mathtools}
\usepackage{graphicx}
\usepackage{float}
\usepackage{nicefrac}       % compact symbols for 1/2, etc.
\usepackage{subcaption}
\usepackage{mathrsfs}
\usepackage{microtype}      % microtypography
\usepackage{multirow}
\usepackage{algorithm}
\usepackage{algpseudocode}
\usepackage{diagbox}
\usepackage{hyperref}       % hyperlinks

\newcommand\FairEditor{FairEdit}
\newcommand{\argmax}{\mathop{\mathrm{argmax}}}
\newcommand{\argmin}{\mathop{\mathrm{argmin}}}

\title{\FairEditor: Preserving Fairness in Graph Neural Networks through Greedy Graph Editing}
\author{Donald Loveland \quad Jiayi Pan \quad Aaresh Farrokh Bhathena \quad Yiyang Lu\\ University of Michigan}
\date{November 2021}

\begin{document}

\maketitle

\begin{abstract}
Graph Neural Networks (GNNs) have proven to excel in predictive modeling tasks where the underlying data is a graph. However, as GNNs are extensively used in human-centered applications, the issue of fairness has arisen. While edge deletion is a common method used to promote fairness in GNNs, it fails to consider when data is inherently missing fair connections. In this work we consider the unexplored method of edge addition, accompanied by deletion, to promote fairness. We propose two model-agnostic algorithms to perform edge editing; a brute force approach and a continuous approximation approach, FairEdit\footnote{Source Code: \url{https://github.com/royull/FairEdit} }. 
\FairEditor ~performs efficient edge editing by leveraging gradient information of a fairness loss to find edges that improve fairness. We find that FairEdit outperforms standard training for many data sets and GNN methods, while performing comparably to many state-of-the-art methods, demonstrating FairEdit's ability to improve fairness across many domains and models. 

\end{abstract}

\section{Introduction}
Over the past decade, deep learning algorithms have become widely used in automated decision-making tasks. Despite neural networks' improved training speed and performance, understanding how neural networks can be adopted to potentially \emph{biased} settings remains an open topic. %Without designing methods to train models appropriately in such conditions, the fair adoption of deep learning remains significantly hindered. 
Discriminatory bias has the potential to appear in many human-centered applications of neural networks, such as in social networks, financial networks, or recommendation systems, where data has been historically generated unfairly. One example of how biased data can manifest comes from mortgage lending in Chicago and Milwaukee over the last five decades. Specifically, redlining (the process of dividing up regions of neighborhoods and restricting access to various services, such as insurance or loans) in racially segregated parts of these cities has caused financial records to disproportionately under-represent various racial and ethnic groups. As a byproduct, some financial institutions have continued to perpetuate these biases by making "data-driven decisions" for lending while failing to recognize the historical context of the data. To this day, data generated through these processes are fed a myriad of decision-making algorithms that have the propensity to amplify any signals present, even if unethical. 

A survey on fairness in machine learning \cite{mehrabi2021survey} identified two reasons for algorithmic unfairness: a) bias in the \emph{data} and b) and an \emph{algorithm}'s susceptibility to bias. They evaluated the \textit{COMPAS} software which measures the risk of an offender recommitting a crime \cite{mehrabi2021survey}, finding heavy racial bias. They cite the unwillingness of the authors to open-source their proprietary data as one of the driving issues. Another example that has demonstrated unfairness is the recent adoption of facial recognition system (FRS). The FRVT 2002 \cite{phillips2003face}, an evaluation of FRS, showed that many algorithms' performance significantly degraded between different genders. Furthermore, several racial biases were also found \cite{givens2003statistical}, highlighting the serious social ramifications that can occur when the systems are used in sensitive scenarios \cite{introna2005disclosive}, such as criminal suspect identification. Based on these sources of unfairness, two possible solutions can be proposed. One solution would focus on facilitating the analysis of bias in data by AI ethics researchers and domain experts. Unfortunately, this can be an arduous and inefficient process. Another route argues for promoting fairness in the decision-making process by using algorithms that are more robust against unfairness. We focus on the latter in this work and consider how one might change the training strategy of a model to promote fairness.

Separately, one new direction of research is developing deep learning algorithms that can operate on graphs. Graphs have been used to represent many real-world systems, such as social networks, transaction networks, and molecule structures. While fairness has been studied for traditional machine learning algorithms and even deep neural networks, graph neural networks (GNNs), neural networks that operate directly on graphs, have only received PUCFYH-YEZMNIminor attention. In this project, we propose to edit the graph adjacency matrix to promote fairness. Intuitively, many cases of historic discrimination have caused data sets to be incomplete due to structural biases that restrict various behaviors. For example, in the context of redlining, different racial and ethnic groups were stratified into distinct communities, causing networks built on this data to be homophily-dominant (i.e. nodes in the local neighborhood all belong to the same sensitive class) with minimal connections across nodes of differing sensitive attributes. Thus, the key assumption that drives our method is that \emph{graph data generated through discriminatory means is either a) missing edges that would have been present in more fair settings, or b) over-representing homophilous edges due to social stratification}. While existing literature proposes different techniques to promote fairness, such as re-weighting existing connections or adversarial-like training, general graph editing methods (including edge addition and deletion) have yet to be explored. Graph editing is an important next step as networks that are locally homophilous, cannot simply be re-weighted to be made fair, as no connections to nodes with differing sensitive attributes exist. Instead, new connections must be added to learn more fair representations. Due to the lack of explainability in GNNs models, learning more fair representations can be one step in improving their applicability in sensitive applications. Likewise, we also intend to learn how different GNN architectures interplay with fairness, helping to elucidate how structure impacts decisions making.

In summary, our contributions in this work are as follows:

\begin{itemize}
    \item We perform a large set of empirical evaluations measuring how fair training mechanisms and models impact various predictive task and fairness metrics, elucidating GNN design choices which improve fairness. 
    \item We propose a new method to introduce fairness into GNN models by editing the graph data during training. The edits either generate new connections, which would have otherwise been there if not for discriminatory factors, or delete old connections which saturate the model. 
    \item We propose a second variant of this method, FairEdit, that relaxes the discrete optimization, required to search the graph edit space, into a continuous optimization problem. Given a set of edits, this approach is able to determine which edit is most impactful to fairness in constant time through gradient approximations. 
    
\end{itemize}

\section{Related Work} \label{Sec:Related work}
 \textbf{Graph Representation Learning} Graph neural networks (GNNs) have been developed as a generalization of convolution neural networks (CNNs) to accommodate graph structured data. Various GNN architectures have been proposed, each introducing a different mechanism to promote expressivity.
 \textbf{GCN} (Graph Convolution Network) \cite{kipf2016semi} proposed a graph representation and propagation rule for convolution neural network to be applied on graph-structured data. Specifically, the work takes the average of all neighbors of a node to update its hidden representations. This can hurt representation learning as proximity usually dominates over useful node features. 
 \textbf{GraphSAGE} (SAmple and aggreGatE) \cite{hamilton2017inductive} worked to generalize GCN to more applications by proposing new aggregation and embedding functions. GraphSAGE operates by concatenating the aggregating the neighborhood embeddings with the hidden representation of the node being updated, allowing for each node to learn unique representations. 
 \textbf{APPNP} (personalized propagation of neural predictions) \cite{klicpera2018predict} recognized the issue of oversmoothing (nodes learn the same representations in deeper GNNs) present in GCN and GraphSage, forcing models to remain shallow. In order to access higher-order networks, the authors adopt a propagation scheme based on personalized PageRank where the model performs a weighted sum between the aggregated neighborhood representations and the representations of the node is updated. 
 
 \textbf{Fairness in Machine Learning} Various notions of fairness in machine learning have been developed in the past decade. The notion of \emph{Individual Fairness}\cite{dwork2012fairness} and \emph{Disparate Treatment} \cite{zafar2017fairness} emphasizes that individuals from different sensitive groups should have similar outcomes if they have similar non-sensitive attributes. \cite{kusner2017counterfactual} introduced \textbf{\emph{counterfactual fairness}}, capturing the intuition that a decision is fair if it is the same in the actual world and in the counterfactual world where the individual belong to a different sensitive group. In contrast, \emph{Group Fairness} concentrates on \textbf{statistical parity}, including \emph{demographic parity (or disparate impact)} \cite{zafar2017fairness}, and \emph{equality of opportunity} \cite{hardt2016equality}. While previous notions focused on prediction outcomes, \emph{Fairness Through Unawareness} \cite{grgic2016case} focused on process fairness, which requires that the sensitive feature are not explicitly included in the decision-making process. Another attempt outside parity in treatment or outcome is a preference-based notion of fairness proposed by \cite{zafar2017parity}, securing parity while guaranteeing high accuracy.

\begin{figure}[h]
    \centering
    \includegraphics[scale=0.25]{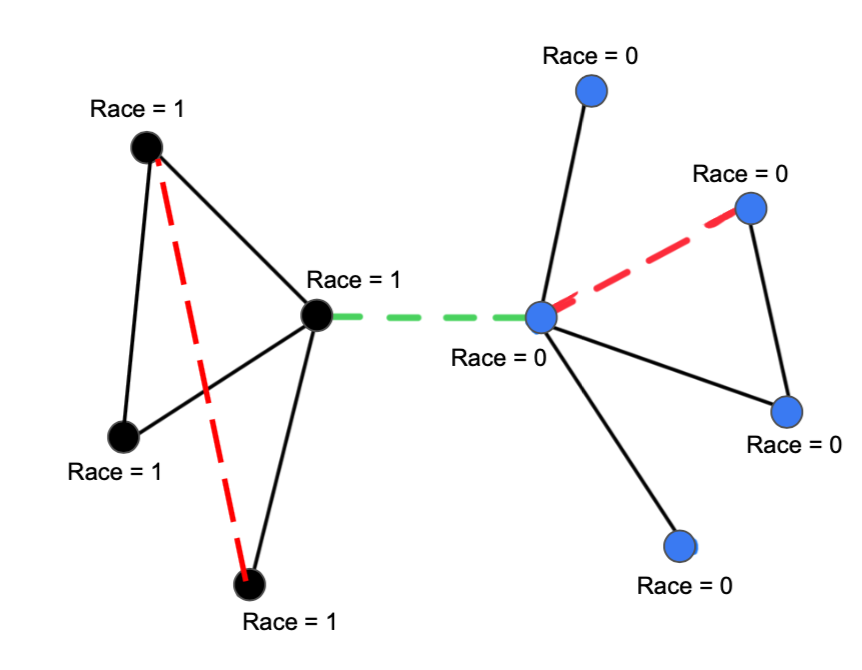}
    \caption{Red dashed edges indicate edges between the same sensitive attribute that have been dropped, and the green dashed edge indicates the edge added between nodes with different sensitive attribute}
    \label{fig:my_label}
\end{figure}

\textbf{Fairness in Graph Representation Learning} The study of fairness in graph-structure data is a very new topic and many domain-specific issues are still open to be addressed.
One major source of bias in graph learning is homophily, which means that similar nodes in graphs tend to interact with each other. \textbf{FairDrop} \cite{FairDrop} proposes to create a random copy of the adjacency matrix biased towards a decrease in homophily and reduce predictability of its sensitive attributes. \textbf{Nifty}\cite{nifty} tries to adopt an adversarial-like training paradigm that perturbs the graph towards a more fair objective. However, this work fails to consider adding an edge when perturbing the graph's structure. \textbf{FairGNN} \cite{dai2021say} considers non-i.i.d. data and tries to leverage graph structure with limited sensitive information, while maintaining high computation efficiency. Node2vec \cite{grover2016node2vec} is an algorithmic framework that learns the mapping of nodes to a low-dimensional vector space with a random walk procedure. \textbf{FairWalk} \cite{fairwalk} extended the node2vec algorithm by grouping neighbors based on sensitive attributes and forcing possible steps to have equal probability with respect to these groups. 
Our method, \textbf{\FairEditor}, will add new connections between vertices that promote fairness, extending upon the work of Nifty and FairDrop. Though \cite{li2020dyadic} claims that "adding fictitious links might mislead the directions of message passing, and further corrupt the representation learning", various works have proven that adding edges can improve various tasks without comprising accuracy \cite{zugner2018adv}.
An illustration is presented in Figure \ref{fig:my_label} to demonstrate what the editing process may look like between two communities of differing sensitive attributes.

\vspace{-.6cm}
\section{Notations and Mathematical Formulations}
 
$G$ is used to denote a graph. A graph $G$ can be represented by a set of nodes $V$ and a set of edges $\mathcal{E}$. We consider $n$ as the number of nodes of graph $G$ which is mathematically given by $n=|V|$. $A \in \{0, 1\}^{n \times n}$ is used to denote the adjacency matrix of $G$, where  $A_{ij} = 1$ indicates an edge between node $i$ and node $j$. Each node in the graph has a set of features $x \in \mathbb{R}^{1\times d}$, where the full set of node features denoted as $\mathbf{X}\in \mathbb{R}^{n\times d}$. The $(.)^{*}$ notation is used to denote an object that has been edited. The specific editing mechanism will be explained when introduced for the first. 

Algorithm and hyper-parameter notations: $\mathcal{F}$ denotes the fairness function and $\mathcal{F}_c$ is used to denote counter factual fairness (explained below). $L$ is used to denote the loss function of the model, the binary cross-entropy loss. The sensitive attribute for the $i^{th}$ node in a graph is given by $x_{i}^s$ and the model parameters are denoted by $\theta$. $\alpha$ is a hyper-parameter of the model which is used to denote the number of edits per iteration and the total number of epochs is given by $K$.

Graph neural networks learn through a message passing framework where node representations are updated based on messages from their neighbors. A simple update for node $v$ at iteration $i$ can be formulated as $X^{i+1}_{v} = F_{l}\big(H(\{X_{u}^{i}|u \in \mathcal{N}(v)\})\big)$ where $F_{l}$ is some embedding function at layer $l$ (e.g. MLP), $H$ is an order agnostic aggregation function (e.g. average), and $\mathcal{N}$ is the set of nodes around a neighborhood of node $v$. Additional learnable functions are sometimes added to improve expressivity of the model, such as an LSTM over the series of learned representations. If the graph has self-loops, the original node $v$ is also included in the neighborhood. At every layer, each set of node features is embedded with the same learned function $F_{l}$, introducing a weight sharing mechanism. At the last layer, $F_{L}$, the learned embeddings are passed to a final classifier or regressor (also usually an MLP) to perform the node prediction task. If the goal is to predict over an entire graph, an additional aggregator is applied to all of the node embeddings before the prediction, resulting in one final output.
%\vspace{-1.5cm}
\section{Methods}
%\vspace{-.5cm}
As mentioned in section \ref{Sec:Related work}, edge re-weighting and edge deletion is a common method used to promote fairness in GNNs. In this work, we consider how one might also include edge addition to introduce relationships otherwise unavailable in the data. We propose two novel approaches that edit a graph's adjacency matrix to determine additions and deletions which improves fairness. The first method searches the set of possible edits within a graph's adjacency matrix, during each training step, and greedily edits based on the edge that provided the largest fairness gain. The second method relaxes the discrete assumption and instead utilizes edge gradients as a signal of importance to fairness. Below, each method is formalized, with the training procedure, and a comparison of time and space complexity.  
%\vspace{-.75cm}
\subsection{Brute Force Edit Method}
%\vspace{-.5cm}
The brute force approach tweaks the underlying graph structure to improve fairness. For a graph $G(V,\mathcal{E})$ we first consider the set of all the possible edge edits on the graph. Let the set of all possible edges of G be $\mathfrak{E}$. Thus a graph of $n$ nodes has $|\mathfrak{E}| =n\times (n-1)$.

At the start of the iteration we make the following change for  $e_i\in \mathfrak{E}$, 
\begin{align} \label{eq:bf edit}
    E_i^*=\begin{cases}
    \mathcal{E}\backslash e_i \qquad & \text{if } e_i\in \mathcal{E} \\
    \mathcal{E}\cup e_i \qquad & \text{if } e_i\notin \mathcal{E}
    \end{cases}
\end{align}
Based on $E_i^*$, the updated edge set in $G^{*}$, the corresponding counterfactual fairness score, $F_C$, (formalized in the evaluation section) is calculated. The procedure is repeated for all $e_i \in \mathfrak{E}$. The final update is made based on the edit that produces the best fairness score. Mathematically, the optimal edited graph $G^*=\argmin_{G^*} F_C(G^{*}) $, i.e. the graph with the edited edge set that maximizes counterfactual fairness. Due to the possibility of creating a graph that is too different from the original graph, we choose $\alpha = 10$. This small number of edits is seen to be generally sufficient while minimizing the computational overhead. 

A major drawback of this approach is its high computational complexity. The run-time complexity of each iteration in this mode is $\mathcal{O}(n^2)$ due to the size of $\mathfrak{E}$, causing issues in the case of large graphs. Due to this drawback, we introduce FairEdit to reduce the run-time complexity.
 
 \begin{algorithm}[H]
\caption{Brute-force Edit}\label{alg:bf}
\textbf{Input:} Training Graph $G=(V, \mathcal{E}, \mathbf{X})$, sensitive index $s_{i}$\\
\textbf{Set:} Number of edits $\alpha$, edit starting epoch $ER$, number of epochs $K$, optimizer $O$, model $\theta_0$ 
\begin{algorithmic}
\State $\theta \leftarrow \theta_0$
\For{$k \leftarrow 1$ $\textbf{to}$ $K$}
    \State Train one step using optimizer $O$ for model parameters $\theta$ on graph $G$
     \If{$k < \alpha$}
        \For{ $e_i \in \mathfrak{E}$ } 
        \State Update $E_i^*$ using equation \ref{eq:bf edit} to obtain $G^{*}$
        \State Calculate $F_C(G^{*})$
        \EndFor
        \State Find optimal edited graph: $G^*=\argmin_{G^*} F_C(G^{*})$
        \State $G \leftarrow G^*$
    \EndIf
\EndFor    
\end{algorithmic}
\textbf{Output:} Optimized model parameters $\theta$ with improved fairness
\end{algorithm}
 
\subsection{FairEdit}

FairEdit retains the same goal as our brute force method of tweaking the underlying graph structure to improve fairness. FairEdit first proposes a set of counterfactual edges $E^*$ to be edited by a) adding edges between nodes with different sensitive attributes with a probability $\rho$ and b) removing edges between nodes with the same sensitive attributes with a probability $\gamma$, creating a new perturbed graph $G^*$. $G^*$ is passed through the GNN model $M$ and the new predictions are compared against the original graph predictions to measure fairness. Given counterfactual fairness is not differentiable, we propose a differentiable approximation of counterfactual fairness through

\begin{equation}\label{eq:Gradient function}
    L = \|M(G) - M(G^{*})\|_{1}
\end{equation}

where a large norm indicates a strong sensitivity in the counterfactual edit. We determine the (approximately) optimal edit by taking the gradient of equation \ref{eq:Gradient function} with respect to $E$ and $E^*$. The magnitude of the gradients can be seen as a relaxation of the fairness increase determined in the brute force method. To choose the final edit, with $\frac{\partial L}{\partial E}$ and $\frac{\partial L}{\partial E^{*}}$, we greedily solve:

\begin{equation}\label{eq:Edge Edit}
    e^{*} = \argmax_{e_{i} \in E^{*}} \begin{cases} \Big|\frac{\partial L}{\partial E_{e_{i}}^{*}}\Big| &\mbox{if $e_{i}$ added to $G^{*}$} \\ \\
\Big|\frac{\partial L}{\partial E_{e_{i}}}\Big| & \mbox{if $e_{i}$ removed from $G$} \end{cases} 
\end{equation}

Intuitively, when an edge is added, we can determine how impactful that change was by looking at the gradient with respect to $E^{*}$. In the deletion cases, impactful edges can be determined by looking at $E$. Note deleted edges will have gradients equal to 0 in $E^{*}$, and added edges will have gradients equal to 0 in $E$. Since the adjacency matrices are discrete, we are unable to directly attribute gradients to each edge. Instead, we learn a scoring matrix $M$ that is able to determine how important an edge is to the loss function. The mask is injected by $A` = A \odot \sigma(M)$, where $\sigma$ is the sigmoid function and $\odot$ indicates element-wise multiplication. $A`$ is discretized into $A^* \in \{0, 1\}^{n \times n}$ by binarizing through a threshold of $0.5$. This process is performed for five iterations, updating the mask at each step to learn an appropriate score. 

By approximating the importance through gradients, the run-time complexity of each iteration in this mode is $\mathcal{O}(1)$ given the need to simply run two forward passes of the model and one backpropagation step. However, this method does require additional space complexity of $\mathcal{O}(n^2)$ as the model requires storage of the additional edges. That said, this space complexity is often much smaller given the probability of addition is usually small.

\begin{algorithm}[H]
\caption{FairEdit}\label{alg:FairEdit}
\textbf{Input:} Training Graph $G=(V, \mathcal{E}, \mathbf{X})$, sensitive index $s_{i}$\\
\textbf{Set:} Number of edits $\alpha$,  number of epochs $K$ 
\begin{algorithmic}
\State $\theta \leftarrow \theta_0$
\For{$k \leftarrow 1$ $\textbf{to}$ $K$}
    \State Train one step using optimizer $O$ for model parameters $\theta$ on graph $G$
    \If{$k < \alpha$}
    \State Generate new graph $G^*$
    \State Calculate $M(G)$ and  $M(G^*)$
    \State Compute approximate gradients for $E$ and $E^*$ from equation \ref{eq:Gradient function}
    \State Solve equation \ref{eq:Edge Edit} to obtain optimal edit $e^{*}$
    \State Perform edit $e^{*}$ on $G$ to get $G^{*}$
    \State $G \leftarrow G^*$
    \EndIf
\EndFor
\end{algorithmic}
\textbf{Output:} Optimized model parameters $\theta$ with improved fairness
\end{algorithm}

\section{Evaluation}

 \subsection{Dataset}
  
 We intend to use three data sets proposed by \cite{nifty} in order to evaluate our GNN models. In each dataset, we will identify the size of graph, the prediction task, and the sensitive attribute we will be analyzing for fairness. Each of our tasks will focus on node classification. All of the data sets are available freely and have been proposed as a standard to benchmark fairness in GNN models. These data sets are useful as they have a well defined sensitive attribute that can be probed for fairness. These datasets include: (a) \textbf{Recidivism Dataset} which is comprised of 18,876 nodes, where each node is representative of a defendant who was released on bail in the U.S state court system during 1990-2009.  Each node contains 18 attributes. Connections are related to similarity of crimes and past convictions. The classification task is to determine whether a defendant would receive bail (i.e., unlikely to commit a violent crime if released) or not (i.e., likely to commit a violent crime) and uses race as the sensitive attribute. (b) \textbf{Credit Default Dataset} is comprised of 30,000 nodes, where each node represents individuals who are utilizing some form of credit. Each node contains 13 attributes. Individuals are connected by their spending and payment behavior. The classification task is to determine whether an individual will default on the credit card payment. Age is used as the sensitive attribute. (c) \textbf{German Credit Dataset} is comprised of 1,000 nodes, where each node represents an individual who uses a specific German bank. Each node contains 27 attributes. Each connection identifies a similarity in credit accounts. The classification task is to classify individuals into those who have high versus low credit risk. The individual’s gender is used as the sensitive attribute.

 \subsection{Measurement metrics}
 
 The model will be evaluated on the three data sets mentioned in the previous section. The predictive performance will be evaluated through the f1-score to mitigate any discrepancies in class distribution, while fairness will be determined by statistical parity, counterfactual fairness, and model stability. Further details of these metrics are below: 
 
 \textbf{Standard Metrics} The standard metrics for binary classification are based on the notation of true positive (TP), true negative (TN), false positive (FP) and false negative (FN). 
 F1-score is defined as the harmonic mean of precision and recall, where precision is defined as true positive divided by all positive predictions and recall is defined as true positive divided by all samples are underlying positive. In mathematical terms, precision is $\frac{TP}{TP+FP}$, recall is  $\frac{TP}{TP+FN}$, and F1-score = $ \frac{Precision*Recall}{Precision+Recall} = \frac{TP}{TP+(FP+FN)/2}$

 \textbf{Fairness Metrics} 
 Counterfactual fairness is measured by calculating the proportion of test nodes whose predicted labels change when the node’s sensitive attribute is flipped. Model stability is measured by calculating the proportion of test nodes whose predicted labels change when test node features are perturbed by a small amount of noise. 
 Statistical parity (SP), also known as group fairness, suggests that the predictor is unbiased if the prediction label \(\hat{Z}\) is independent of the protected sensitive attribute S and can be computed by \textbf{\(\Delta_{SP}\)} = \(\big|Pr(\hat{Z}=1|S=1)-Pr(\hat{Z}=1|S=0)\big|\). Equal opportunity (EO) assumes that similar nodes with different sensitive attributes should have similar outcomes. This is computed as \(\Delta_{EO}=\big|Pr(\hat{Z}_u=1|Z_u=1,S=1)-Pr(\hat{Z}_u=1|Z_u=1,S=0)\big|\).
 
 \subsection{Evaluation Process and Results}

We evaluate our proposed fairness-training framework on a variety of modern graph representation learning and GNN architectures such as GCN \cite{kipf2016semi}, GraphSage \cite{hamilton2017inductive} and APPNP \cite{klicpera2018predict}. Each model is also applied to other state-of-the-art fair training frameworks such as FairGNN \cite{dai2021say} and Nifty \cite{nifty} to compare against. The evaluation metrics mentioned in 5.2 are calculated for the three data sets identified in 5.1. We perform hyper-parameter searches for each dataset, model, and training method, tuning the learning rate ($lr \in \{10^{-3}, 10^{-4}, 10^{-5}\}$), hidden size ($h \in \{16, 32\}$), and model depth ($l \in \{2, 3\}$). Additionally, we compare against FairWalk \cite{fairwalk}, a non-GNN fair model baseline that operates on graphs. Given FairWalk only produces node embeddings, a random forest model is used for the actual classification.  

The results for the proposed evaluation metrics are shown in table 1. Results for the APPNP model and FairGNN training method are not provided given the APPNP model does not adopt the required architecture for FairGNN. Likewise, results for the brute force method are not provided for the credit defaulter dataset due to its exceedingly high computation cost. These issues necessitate methods such as FairEdit which are model agnostic and do not scale with the graph size. For methods previously proposed, such as FairGNN and Nifty, we re-implement their code and repeat the authors' analysis. Unfortunately, even with reported hyperparameters, we are unable to match the results presented in the previous papers. Thus, in our results table and discussion, we consider our results.  

For many datasets and models, FairEdit outperforms standard training, usually improving F1-score and fairness. Notably, FairEdit applied to the Recidivism dataset outperforms the standard training baseline on every model and often outperforms FairGNN and Nifty both in predictive performance and fairness. That said, for the German credit dataset and Credit defaulter dataset, FairGNN, Nifty, and FairEdit each demonstrate varying performance depending on the model choice. This insight is important as studies comparing FairGNN and Nifty are nonexistent, and their variability indicates more work needs to be to create more stable fairness methods. In addition to FairEdit's performance, we are also able to better understand how model architecture impacts performance. In the recidivism and credit datasets, GraphSage seems to perform best in both F1-score and fairness, arguing that the concatenation between node features and neighbor features may introduce a mechanism that promotes fairness. Furthermore, despite FairEdit only approximately solving for the best edge to edit, result comparisons between brute force and FairEdit indicate this approximation is enough to sufficiently improve fairness while maintaining performance. 

Despite the apparent success in a handful of scenarios, FairEdit has instances where it doesn't perform well, particularly on the German credit and credit defaulter datasets. Interestingly, when we look into the percentage of sensitive attributed nodes in all of the datasets analyzed, we find that they are evenly distributed across the two classes. In other words, the datasets are inherently very fair to begin with and do not display any proportionality issues. This is important as we initially assumed the data experienced some sort of bias between the sensitive attribute and class distribution. This issue speaks broadly to the lack of realistic datasets which mimic biased behavior seen in the real world and may partially explain our mixed performance. 

\begin{table*}[!ht]
\centering
\caption{
Fair Training Results: Arrows ($\uparrow$, $\downarrow$) indicate the direction of better performance. BF indicates brute-force algorithm
}
\label{tab:group_acc}
% \setlength{\tabcolsep}{5pt}
% \vspace{-2mm}
{\small \begin{tabular}{clccccc}
Dataset & Method &  F1-score ($\uparrow$) & Unfairness ($\downarrow$) & Instability ($\downarrow$) & $\Delta_{SP} (\downarrow$) & $\Delta_{EO} (\downarrow$) \\
\toprule
\multirow{15}{2cm}{German credit graph}
&Fairwalk&0.790&0.400&0.360&0.019&0.009\\\cline{2-7}
&GCN& 0.646 & 0.148 & 0.160 & 0.406 & 0.379\\
% above: dp: 0.4 hidden: 64 lr 1e-3 ep 300 seed 1
&FairGCN& 0.810 & 0.024 & 0.024 & 0.109 & 0.023\\
% above: dp: 0.8 hidden: 8 num of output layer: 4  lr 1e-4 ep 150 seed 1
&Nifty-GCN& 0.740 & 0.016 & 0.092 & 0.448 & 0.368\\
&GCN-BF& 0.646 & 0.156 & 0.168 & 0.428 & 0.409\\
% above: dp:0.7 hidden: 32 lr 1e-3 ep 150 seed 1
&GCN-FairEdit& 0.692 & 0.052 & 0.168 & 0.448 & 0.463\\\cline{2-7}
&SAGE& 0.749 & 0.224 & 0.304 & 0.355 & 0.303\\
% above: dp:0.8 hidden: 64 lr 1e-3 ep 150 seed 1
&FairSAGE& 0.748 & <0.001 & <0.001 & 0.003 & 0.055\\
% above: dp: 0.5 hidden:16 num_layer:32   lr 1e-3 ep 150 seed 1
&Nifty-SAGE& 0.737 & 0.036 & 0.132 & 0.659 & 0.554\\
&SAGE-BF& 0.749 & 0.188 & 0.316 & 0.416 & 0.363\\
% above: dp:0.8 hidden: 64 lr 1e-3 ep 150 seed 1
&SAGE-FairEdit& 0.684 & 0.219 & 0.324 & 0.347 & 0.319\\ \cline{2-7}
&APPNP& 0.772 & 0.092 & 0.436 & 0.241 & 0.180\\ 
% above: dp:0.4 hidden: 16 lr 1e-4 ep 300 seed 1
&Nifty-APPNP& 0.762 & 0.038 & 0.142 & 0.542 & 0.523\\
&APPNP-BF& 0.772 & 0.088 & 0.436 & 0.241 & 0.180\\
% above: dp:0.4 hidden: 16 lr 1e-4 ep 300 seed 1
&APPNP-FairEdit& 0.734 & 0.184 & 0.124 & 0.343 & 0.288\\
 \midrule

\multirow{15}{2cm}{Recidivism graph}
&Fairwalk&0.419&0.483&0.482&0.009&0.004\\\cline{2-7}
&GCN& 0.773 & 0.085 & 0.226 & 0.099 & 0.046\\
&FairGCN& 0.727 & 0.064 & 0.153 & 0.071 & 0.067\\
% above: dp: 0.5 hidden: 32 number of layer: 16  lr 1e-3 ep 150 seed 1
&Nifty-GCN& 0.663 & 0.015 & 0.123 & 0.025 & 0.028\\
&GCN-BF& 0.779 & 0.091 & 0.427 & 0.004 & 0.018\\
&GCN-FairEdit& 0.813 & 0.004 & 0.175 & 0.078 & 0.014 \\ \cline{2-7}
&SAGE& 0.778 & 0.087 & 0.430 & 0.002 & 0.018\\ 
&FairSAGE& 0.810 & 0.056 & 0.472 & 0.012 & 0.018\\ 
% above: dp: 0.5 hidden:16 num_layer:32   lr 1e-3 ep 150 seed 1
&Nifty-SAGE& 0.834 & 0.004 & 0.233 & 0.058 & 0.034\\
&SAGE-BF& 0.779 & 0.092 & 0.427 & 0.004 & 0.018\\
&SAGE-FairEdit& 0.823 & 0.053 & 0.475 & 0.015 & 0.032\\ \cline{2-7}
&APPNP& 0.747 & 0.015 & 0.168 & 0.080 & 0.057\\ 
% above: dp:0.4 hidden: 32 lr 1e-3 ep: 200 seed 1
&Nifty-APPNP& 0.752 & 0.011 & 0.174 & 0.094 & 0.042\\
&APPNP-BF& 0.749 & 0.015 & 0.168 & 0.083 & 0.062\\
&APPNP-FairEdit& 0.759 & 0.011 & 0.135 & 0.063 & 0.034\\
 \midrule

\multirow{12}{2cm}{Credit defaulter graph}
&Fairwalk&0.722&0.412&0.540&0.008&0.001\\ \cline{2-7}
&GCN& 0.793 & 0.162 & 0.282 & 0.137 & 0.136\\
&FairGCN& 0.739 & 0.052 & 0.245 & 0.045 & 0.051\\
% above: dp: 0.5 hidden: 4 num of layer: 4  lr 1e-4 ep 150 seed 1
&Nifty-GCN& 0.799 & <0.001 & 0.136 & 0.110 & 0.093\\
&GCN-FairEdit& 0.797 & 0.123 & 0.195 & 0.122 & 0.125\\ \cline{2-7}
&SAGE& 0.820 & 0.130 & 0.378 & 0.092 & 0.094\\ 
&FairSAGE& 0.825 & 0.061 & 0.382 & 0.125 & 0.097\\ 
% above: dp: 0.5 hidden: 32 num of layer: 16  lr 1e-3 ep 150 seed 1
&Nifty-SAGE& 0.833 & 0.006 & 0.134 & 0.112 & 0.090\\ 
&SAGE-FairEdit& 0.796 & 0.173 & 0.326 & 0.088 & 0.094\\ \cline{2-7}
&APPNP& 0.824 & 0.065 & 0.141 & 0.113 & 0.109\\ 
% above: dp:0.5 hidden: 16 lr 1e-3 ep 150 seed 1
&Nifty-APPNP& 0.084 & 0.032 & 0.013 & 0.012 & 0.011 \\
&APPNP-FairEdit& 0.823 & 0.063 & 0.136 & 0.110 & 0.104\\
\bottomrule
\end{tabular}}
\end{table*}

\section{Conclusion}

In this work we propose a novel training method to promote fairness in GNNs through graph editing, i.e., adding or deleting edges of a graph. Given the complexities involved with solving the originally proposed optimization problem, we also develop a relaxed version with significantly improved run-time. The brute force technique showed positive results for the smaller data sets on which it was implemented, but failed to scale to larger graphs. FairEdit, the relaxed variant, proved able to compete with state-of-the-art fairness techniques such as FairGNN and Nifty on many experiments, while maintaining model agnosticism. Furthermore, our experiments were able to help us elucidate GNN architectures that intrinsically promote fairness; in a significant number of experiments, GraphSAGE was able to outperform other models independent of the training method.

Our results indicate that empowering graph editing, specifically with the inclusion of edge additions, can improve GNN fairness without comprising accuracy. In the future, we hope to improve upon this work both at an experimentation and algorithmic design level. For data, we would find new datasets which demonstrate significant bias and disparity, similar to what can be found in real world applications, in order to sufficiently test both our models and other models. We hypothesize that other models which do not specifically handle our assumptions regarding missing and over-represented edges may not perform as well in these scenarios. As for method development, we believe more work can be done to improve the quick and efficient optimization of graph edits. Currently, the gradient-based method experiences a significant amount of noise, as seen in fields such as explainable AI, meaning the edge gradients are likely often not optimal. Together, these two scenarios can demonstrate the power of the FairEdit algorithm while improving convergence and expressivity.

\clearpage
\bibliography{main}

\end{document}